\documentclass[accepted]{uai2023} 

\usepackage[american]{babel}

\usepackage{natbib} 
\usepackage{mathtools} 
\usepackage{booktabs} 
\usepackage{tikz} 
\usepackage{graphicx}
\usepackage{subcaption}
\usepackage{amsfonts}

\title{Defensive Perception: Estimation and Monitoring of Neural Network Performance under Deployment}

\author[1]{\href{mailto:<hendrik.vogt@zf.com>?Subject=Defensive Perception}{Hendrik Vogt}{}}
\author[1]{Stefan Buehler}
\author[1,2]{Mark Schutera}
\affil[1]{%
    ZF Friedrichshafen AG\\
    Friedrichshafen, Germany\\
    @zf.com
}
\affil[2]{%
    Karlsruhe Institute of Technology, Karlsruhe, Germany, @kit.edu
  }
  
\begin{document}
\maketitle

\begin{abstract}
In this paper, we propose a method for addressing the issue of unnoticed catastrophic deployment and domain shift in neural networks for semantic segmentation in autonomous driving. Our approach is based on the idea that deep learning-based perception for autonomous driving is uncertain and best represented as a probability distribution. As autonomous vehicles' safety is paramount, it is crucial for perception systems to recognize when the vehicle is leaving its operational design domain, anticipate hazardous uncertainty, and reduce the performance of the perception system. To address this, we propose to encapsulate the neural network under deployment within an uncertainty estimation envelope that is based on the epistemic uncertainty estimation through the Monte Carlo Dropout approach. This approach does not require modification of the deployed neural network and guarantees expected model performance. Our \textit{defensive perception envelope} has the capability to estimate a neural network's performance, enabling monitoring and notification of entering domains of reduced neural network performance under deployment. Furthermore, our envelope is extended by novel methods to improve the application in deployment settings, including reducing compute expenses and confining estimation noise. Finally, we demonstrate the applicability of our method for multiple different potential deployment shifts relevant to autonomous driving, such as transitions into the night, rainy, or snowy domain. Overall, our approach shows great potential for application in deployment settings and enables operational design domain recognition via uncertainty, which allows for defensive perception, safe state triggers, warning notifications, and feedback for testing or development and adaptation of the perception stack.
\end{abstract}

\section{Introduction}
\label{sec:intro}

The ability to accurately perceive semantic information is critical in autonomous driving and automotive vision. Neural networks can perform this task through semantic segmentation, where a neural network assigns a class label to each pixel of an input image \citep{xue2018survey}. State-of-the-art semantic segmentation models have achieved high performance across a wide range of scenarios and domains \citep{behley2019iccv,bdd100k}.\\

Pixel-wise semantic segmentation is the task of assigning every pixel of the input image a class prediction. Because every pixel is classified with a value, semantic segmentation is capable of drawing fine outlines, generating high-semantic information central to modern perception within autonomous driving. In this work, we utilize DeeplabV3+~\citep{DBLP:journals/corr/abs-1802-02611} as a base model for pixel-wise semantic segmentation and modify it for use with Monte Carlo Dropout. By inserting dropout layers \citep{gal2016dropout} after every 2D convolution layer, we demonstrate that any model can be modified to perform Monte Carlo Dropout, even after training.\\

Deploying neural networks as part of the perception system for an autonomous vehicle requires a clear understanding of the operational design domain (ODD) in which the vehicle will operate. Therefore, the ODD is defined explicitly, and data is gathered and used to optimize and validate the neural network for this specific domain. However, in real-world applications, deep convolutional neural networks (CNNs) may be exposed to substantially different data from the training data, leading to a phenomenon known as deployment shift \citep{chan2021entropy, SCHUTERAdomainessence.2019}. This deployment shift can occur due to various reasons like a change in time of day, weather, landmarks, object appearance, and traffic conditions, making it impossible to consider every possible scenario, use case, and road condition while defining the ODD \citep{Koopman1}. 
The limitations of ODD definitions in ensuring safety in autonomous driving are highlighted by the effect that
it cannot cover every possibility of change in the real world, which may cause severe security risk and fatalities \citep{BANKS2018278, hullermeier2021aleatoric, nhtsaCrashes2022}. Hence, there is an emerging need for autonomous systems to recognize and understand when they are in unknown and potentially unsafe situations, especially during inference when the system has been deployed. Safety-inducing strategies that detect unknown situations and transition the system into a safe state referred to as defensive perception are required. The perception system needs to have a mechanism to detect when it is leaving the ODD and anticipate hazardous uncertainty, which reduces the performance of the perception system.

As previously stated, deployment shift, the phenomenon of a neural network being exposed to data from a different domain than its training data, can occur in real-world applications of autonomous vehicles. However, detecting deployment shift is difficult due to the lack of correlation between deployment shift and drop in prediction confidence, as highlighted by Nguyen et al. in \citep{nguyen2015deep}.

Several state-of-the-art approaches have been proposed to address this issue of out-of-domain detection and uncertainty estimation in deep neural networks as discussed in \citep{gawlikowski2021survey}. For instance, Du et al. proposed Virtual Outlier Synthesis (VOS) \citep{du2022vos}, a method that synthesizes outliers for additional training to generate a clear boundary between in- and out-of-domain data. Another approach \citep{chan2021entropy} shows that by retraining a semantic segmentation model on unknown objects to maximize the prediction's softmax entropy, the uncertainty of specific out-of-domain object instances can be detected. Additionally, deploying auxiliary models such as an Essence Neural Network \citep{blazek2021explainable}, a Posterior Neural Network \citep{charpentier2020posterior}, or Meta Classifiers \citep{rottmann1811prediction} enables the estimation of domain affiliation or uncertainty of a sample's prediction. 

While these approaches may require additional models, architectural adaptions \citep{UncertaintySoftmaxToReLU2018} or dedicated training processes \citep{uncertaintyEstimationTrainingMethod2020}, another alternative is to use minimally invasive approaches such as Monte Carlo Dropout \citep{rottmann2019uncertainty} for estimating epistemic uncertainty in deep neural networks.\\

Conventional neural networks struggle to express prediction confidence, especially when leaving the source domain they have been trained on. 

As dropout at first was introduced to prevent a neural network from overfitting and thus was only applied at training to generalize the model's prediction \citep{DropoutHinton2012} especially Monte Carlo Dropout later was introduced as a method to measure the model's uncertainty of non-Bayesian neural networks by applying the Monte Carlo Dropout also during inference and determine its predictive distribution \citep{gal2016dropout}. 

Monte Carlo Dropout mimics multiple sub-network prediction distributions $q(\textbf{y}|\textbf{x})$ by deploying dropout layers throughout the complete network and multiple forward passes $T$. 
While $W_i$ denotes the networks weight matrices and $L$ enumerates the model's layer. The deviations in the sub-networks' predictions $\hat{\textbf{y}}$ are then utilized to express the epistemic uncertainty of the entire model on a single frame \textbf{x} - referred to as Monte Carlo Dropout, giving the estimated approximate predictive distribution:

\begin{equation}
    \mathbb{E}_{q(\textbf{y}|\textbf{x})}(\textbf{y}) \approx \frac{1}{T}\sum^T_{t=1} \hat{\textbf{y}} (\textbf{x}, \textbf{W}_1^t,  \dots, \textbf{W}_L^t).
\end{equation}
\label{eq_mcd}\\

In this paper, we propose a method for estimating epistemic uncertainty during the inference of semantic neural networks for autonomous driving using Monte Carlo Dropout,
as the widely accepted uncertainty measurement at inference \citep{MCDropoutUncertainty2019}.

In contrast to the other state of the art uncertainty measurements mentioned above using other techniques than the Monte Carlo Dropout, the here presented approach aims to be applicable on any neural network at inference and in real time without making changes at training and or test time as well as it does not require additional data. 

Incorporation into our \textit{defensive perception envelope}, which monitors uncertainty during deployment, demonstrates that epistemic uncertainty can serve as a proxy for model performance. This novel approach allows us to detect when the system is operating outside of its intended domain and provides an online cue for prediction performance. Furthermore, by imposing thresholds on the uncertainty value, we can define triggers that can be used to implement safety measures such as warning notifications for the driver or even transitions into a \textit{safe state} where the vehicle engages other safety systems and for example, reduces its speed. The main contributions of this paper are:
\begin{itemize}
    \item A safety envelope that integrates Monte Carlo Dropout \citep{MCDropoutUncertainty2019} into semantic segmentation for autonomous driving scenarios.
    \item A novel entropy measure that captures model performance and domain shifts during deployment and inference.
    \item An adaptation of the Monte Carlo Dropout method that utilizes rolling forward passes to improve computational efficiency during deployment.
\end{itemize}

\section{Novel Methods and Metrics}

\subsection{Novel concept for defensive perception}

\begin{figure}[h]
  \centering
  \includegraphics[width=0.94\linewidth]{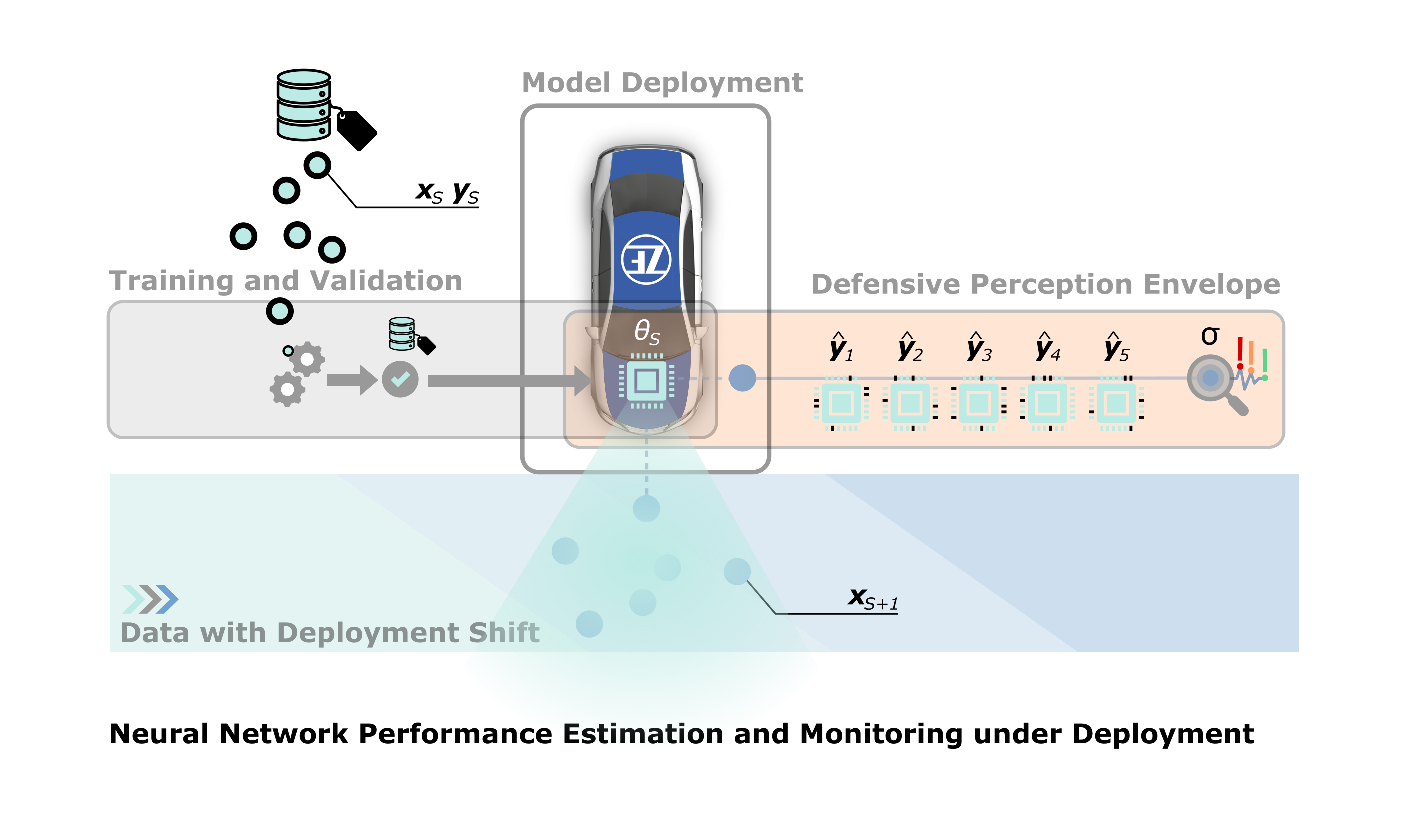}
  \caption{\textbf{System Flow Overview} - within an autonomous driving platform supported by a perception stack, the perception model $\theta_S$ is trained and validated for a given source domain $S$. Deployed in the vehicle the \textit{defensive perception envelope} generates five outputs $\mathcal{Y} = \{ \hat{\textbf{y}}_1, \dots, \hat{\textbf{y}}_5\}$ with Monte Carlo Dropout. Suppose a sample with a domain shift $\textbf{x}_{S+1}$ is fed to the perception model $\theta_S$, the uncertainty of the output vectors for this sample rises, and the \textit{defensive perception envelope} informs the system that it enters a domain with high uncertainty. These notifications are triggered by a pre-selected threshold ~$\sigma$.
  }
  \label{fig:Framework}
\end{figure}

In the following, we present a detailed overview of our proposed method for estimating uncertainty and monitoring the performance of neural networks during deployment. Our approach utilizes a \textit{defensive perception envelope}, which is wrapped around a given perception algorithm. Typically, the performance of a neural network is evaluated by comparing its predictions to manually labeled data (ground truth). However, such labeled data is unavailable during online inference in autonomous driving. To address this, our \textit{defensive perception envelope} indirectly estimates the neural network's performance using Monte Carlo Dropout, enabling real-time performance estimation during deployment. Fig.~\ref{fig:Framework} illustrates the schematic of our proposed framework. \\

\textbf{Training and Validation}
The base of our approach is a perception neural network $\theta_S$ that solves a task such as pixel-wise semantic segmentation. The neural network is therefore trained, validated, and released for deployment in the source domain $S$ with samples $\textbf{x}_S$ from the said domain.\\

\textbf{Model Deployment}
For deployment, the model is modified by inserting dropout layers \citep{gal2016dropout} after every 2D convolution layer. The modification prepares the neural network for the Monte Carlo Dropout approach of our \textit{defensive perception envelope}.\\

\textbf{Data with Deployment Shift}
During deployment, due to a domain shift, the neural network is prone to encounter samples $\textbf{x}_{S+1}$, which are outside of the source domain. There are numerous reasons for a domain shift, including deployment shifts and other ODD shifts that have not been considered during training and validation. The emerging shifts and the resulting potential drop in performance are critical as they occur silently and result in unnoticed catastrophic deployment.\\

\textbf{Defensive Perception Envelope}
Monte Carlo Dropout determines the uncertainty $u_t$ for a sample $\textbf{x}_t$ at time $t$. The uncertainty is calculated by multiple forward passes $n$ of the same sample while randomly dropping different weights. The fluctuation in the predictions $\hat{\textbf{y}}_t \in \{\hat{\textbf{y}}_{t,1}, \dots, \hat{\textbf{y}}_{t,n}\}$ is mapped to an uncertainty metric (see Subsec.~\ref{subsec:subsec_metric})
. The \textit{defensive perception envelope} is configured permissive or stringent by introducing uncertainty thresholds based on the system characteristics. The uncertainty threshold can be based on the uncertainty value distribution computed on the overall data. Further, multiple thresholds enable triggers for multiple stages of defensive reactions, such as system notification, vehicle slow-down, and the safe-state transition.

\subsection{Pseudo Cross-Entropy for uncertainty during inference}
\label{subsec:subsec_metric}

At the core of our uncertainty metric resides the cross-entropy $CE$ metric. As input, the cross-entropy expects a probability distribution $\textbf{q}$ of the prediction vector $\hat{\textbf{y}}$, in the form of a normalized exponential function over all predictable classes $c \in \textbf{C}$, such as given by a softmax layer,
\begin{equation}
    \textbf{q}_c = \frac{
                        e^{\hat{\textbf{y}}_{c}}
                        }
                        {
                        \sum_{i}^{\textbf{C}} e^{\hat{\textbf{y}}_{i}}
                        }.
\end{equation}

The entropy $\textbf{H}$ of a prediction vector $\textbf{q}$ is calculated by multiplication with the true distribution $\textbf{p}$. During deployment, a true distribution is not given; thus, the approach makes use of a pseudo ground truth approximation $\textbf{p}_i \approx \tilde{\textbf{y}}'_i$,

\begin{equation}
    H(\textbf{p},\textbf{q}) = - \sum_{i}^{\textbf{C}}\tilde{\textbf{y}}'_i~log(\textbf{q}_i).
\end{equation}
\label{eq:entropy}

Assuming that a true prediction is linked with a single class, the pseudo ground truth is approximated as a one-hot-encoding of the prediction vector $\textbf{q}$, resulting in the pseudo cross-entropy $CE'$,
 
\begin{equation}
    CE' = - log(\frac{e^{\hat{\textbf{y}}_c}}{\sum_{i}^{\textbf{C}}e^{\hat{\textbf{y}}_i}}).
\end{equation}

In order to deploy the pseudo cross-entropy as an uncertainty measure of a neural network's prediction, further requirements need to be fulfilled:
\begin{itemize}
    \item The pseudo cross-entropy needs to depict the entropy emerging from multiple forward passes $T$.
    \item The entropy needs to be independent of the number of forward passes $n$.
    \item Entropy should follow an exponential function to smooth uncertainty for small deviations while upscaling larger deviations.
    \item For comparability, the range of values needs to be confined to $CE' \in [0,1]$.
\end{itemize}


To measure the uncertainty over multiple forward passes, the one-hot encoded output vector $\textbf{y}_{fwp}$ from each forward pass ($fwp$) is taken, and the hits for each class are accumulated. 
The retrieved vector $\textbf{v}_{hC}$ shows the distribution of hits over all classes for the number of applied forward passes. As only the classes, which are predicted in at least one of the forward passes, are of interest, any class with $0$ hits is removed from this vector so that  $\textbf{v}_{hC} \setminus \{0\}$, and respective a class $i$ is represented by $\mathcal{C}_h$.

From this vector, the class with the maximum number of hits is assumed to be the true class, hence a pseudo ground truth $max(\textbf{v}_{hC})$ (equal to $\textbf{p}$ in Eq.~\ref{eq:entropy}) is determined. In the case of two or more classes having the maximum number of hits, the class with the lowest index is taken as the pseudo ground truth, following the implementation of the used $argmax$ function, provided by the python library NumPy \citep{Numpy}.

The here presented formula fulfilling the requirements mentioned above is:
\begin{equation}
    CE_u = 1-\frac{\exp({\frac{max(\textbf{v}_{hC})}{n_{fwp}}})}{\sum_{i}^{\mathcal{C}_h} \exp({\frac{\textbf{v}_{hC}(i)}{n_{fwp}}})}.
\end{equation}

For intuitive readability, the uncertainty measurement is subtracted from one to have a low score near zero when the uncertainty of the neural network is low and a value near one when the uncertainty is critical. For semantic segmentation, the classification is done pixel-wise; hence the uncertainty is calculated on every pixel of the given input frame. In order to obtain the frame's overall uncertainty, the mean of all pixel's uncertainties is determined.

\subsection{Rolling Monte Carlo Dropout}\label{subsec:RollingMC}

The frame rate of advanced driver assistance systems (ADAS) or autonomous vehicle perception systems must be high enough to allow the system to react on time to its surrounding environment. As a result, consecutive frames in a sequence $\mathcal{S}$ tend to be similar (see Subsec.~\ref{subsec: experiment rolling MC}). In order to reduce computational effort and increase the efficiency of the implemented \textit{safety envelope}, we introduce the Rolling Monte Carlo Dropout method.

This method is based on the idea of a sliding window over the sequence $\mathcal{S}$. Instead of applying Monte Carlo Dropout on a single image multiple times (shown in Fig.~\ref{fig:Framework}), the Monte Carlo Dropout is applied to a sequence of consecutive images. Sequential data allows the calculation of the modified categorical cross entropy $u_t$ for the Rolling Monte Carlo Dropout by replacing the number of forward passes $n_{fwp}$ with the number of images $n_{img}$ within the stride of the defined sliding window.

\begin{equation}
    \mathcal{S}_t = \{ \textbf{x}_{t-n} \dots, \textbf{x}_{t}\}
\end{equation}

\begin{equation}\label{Eq:SlidingWindow}
    u_{t} = \sum_{\textbf{x} \in \mathcal{S}_t} CE_u(\textbf{x})
\end{equation}

The uncertainty of our measurement increases when applying the Rolling Monte Carlo Dropout method to a sequence of consecutive images rather than a single image due to the induced aleatoric uncertainty. Hence, the Rolling Monte Carlo Dropout method cannot be applied to an arbitrary number of consecutive images. Instead, the number $n$ of images in a sequence or window is constrained by the speed range of the ego vehicle and the sensor's sampling rate, influencing the magnitude of the aleatoric uncertainty. 
Overall, using the Rolling Monte Carlo Dropout improves the efficiency of the \textit{defensive perception envelope} by reducing the required number of forward passes while maintaining the model's accuracy.
The operational capabilities and boundaries are the subject of study in the following experiments (see Sec.~\ref{sec:sec_Experiments}).

\section{Experiments}
\label{sec:sec_Experiments}

In this study, we conduct experiments using two datasets: the MNIST dataset and the BDD10K dataset. The MNIST dataset, representing a simple machine learning task, is used to provide a general proof of concept for the proposed metric. The BDD10K dataset, represents a more complex task - semantic segmentation - and serves as a real-world example in the context of autonomous driving.\\

\textbf{MNIST} \citep{lecun2010mnist} consists of 70,000 handwritten numbers containing the digits zero through nine and accordingly labeled, suitable for a classification task. MNIST is deployed as a toy problem, being well-known, easy to interpret, and comprehensible.
\\

\textbf{BDD10K} \citep{bdd100k} contains 9.000 images labeled pixel-wise for semantic segmentation. The images are mostly non-sequential and suitable for single-frame prediction only. The dataset is chosen for its diversity, including images from different times of day, weather conditions, and scenarios. This allows for domain shift experiments. Further, there are unlabeled video sequences that can be used for deployment and runtime experiments.\\

The models for each experiment are only trained on a defined source domain: for the MNIST dataset, this is non-rotated numbers, and for the BDD10K dataset, it is images recorded during the day or labeled as clear. Any data that is not part of the source domain is framed and subsequently interpreted as a domain shift. This data is then only used at inference time for validation and test purposes.\\

The experiments, particularly those related to semantic segmentation, are designed to demonstrate the validity of the proposed metrics and the methods underlying the \textit{defensive perception envelope} (see Fig.~\ref{fig:Framework}. To apply the proposed approach, the neural network must be trained on explicit classes and should not contain any general unknown or misc class, as proposed in \citep{zhang2017universum}. The base dataset for these experiments is the BDD10K dataset, and the task is semantic segmentation. Training is conducted on NVIDIA P100 GPUs with Intel Xeon E5-2667 v4 CPUs. For details on neural network architecture, implementation, and tooling, see our repository\footnote{Defensive Perception Repository (ours): \url{https://osf.io/fjxw3/}}, including a jupyter notebook kick-start demo.\\

\subsection{How confident should we be about confidence?}
\label{subsec: confidence_lie}
In the first experiment, we introduce a domain shift to the MNIST dataset by anticlockwise rotating the given samples in five-degree increments up to a total of 90 degrees. The classifier, which is only trained on the source domain (non-rotated samples), is equipped with dropout layers. The uncertainty (as defined in Subsec.~\ref{subsec:subsec_metric})
of each domain is calculated using Monte Carlo Dropout on 20 forward passes for each sample. Since the labels and correct class for the out-of-domain samples are available, we compare the uncertainty and performance estimation to the prediction error and model confidence derived from the maximum one-hot encoded output vector of the model's prediction.

\begin{figure}[h]
  \centering
  \includegraphics[width=0.45\textwidth]{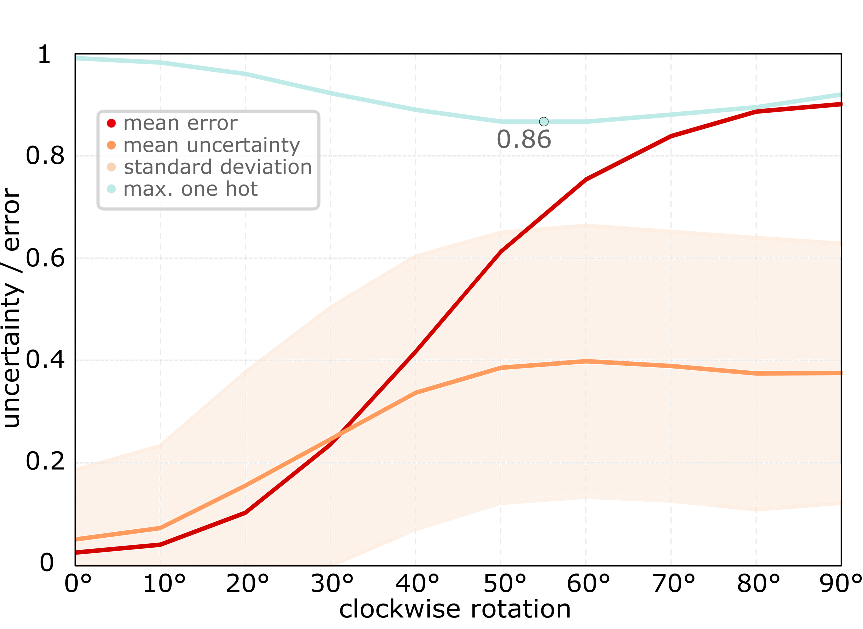}
  \caption{This graph shows the uncertainty on the MNIST test data anticlockwise rotated up to $90^{\circ}$, computed with 20 forward passes and a dropout rate of 0.4. The model was trained on the MNIST \citep{lecun2010mnist} training data set without any applied rotation.}
  \label{fig:Mnist_Finding0_CC}
\end{figure}

Both the error and uncertainty increase with the rotation (see Fig.~\ref{fig:Mnist_Finding0_CC}). 
It is worth noting that even though the model's performance on the out-of-domain data drops by over 90\%, its confidence merely drops by 14\%. 


This behavior is substantiated by Spearman's rank correlation coefficients \citep{Spearman_corr}. While the uncertainty and the model error have a correlation coefficient of $0.93$, the correlation coefficient of the maximum one hot encoded values with respect to the error is nine percentage points lower.



 This supports the findings of Nguyen et al. \citep{nguyen2015deep}, stating that a model's confidence does not reliably detect out-of-domain data. On the contrary, our proposed uncertainty metric provides a reliable performance estimation that reflects the model's uncertainty, as evidenced by its strong correlation with the model's error. This is an important finding, as in real-world deployment scenarios, labels are typically not provided during inference, making it difficult to determine a model's performance by means of conventional offline validation.

\subsection{Uncertainty is able to depict out-of-domain performance}

\begin{figure*}
    \centering
    \begin{subfigure}[a]{1\textwidth}
        \centering
        \includegraphics[width=\textwidth]{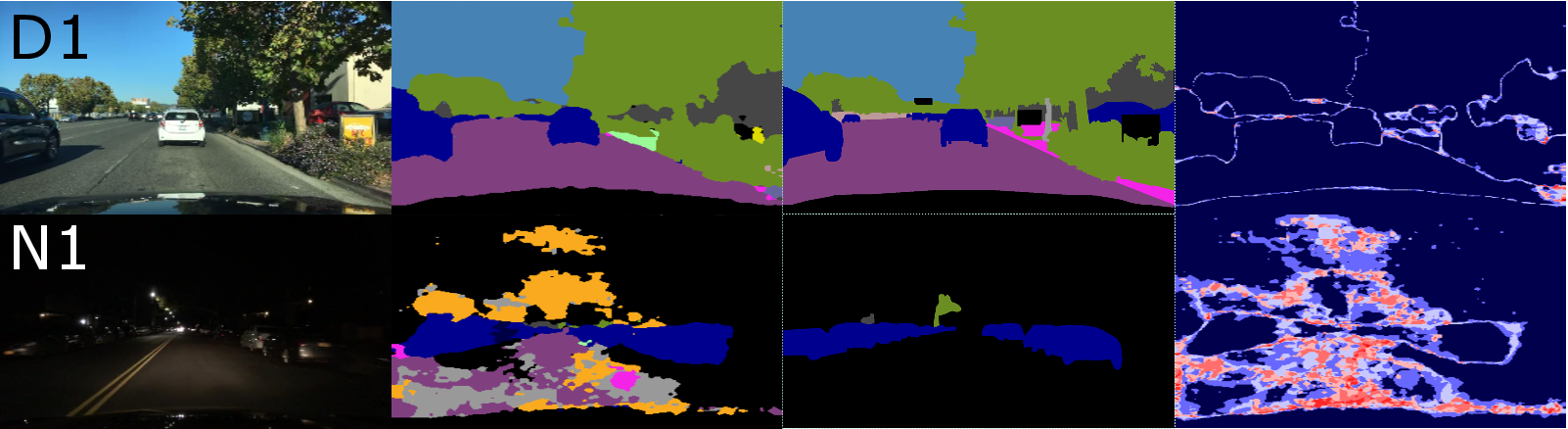}
        \caption{The first row shows the day domain (D1) and the second row the night domain (N1). From left to right: Raw camera image, model's prediction, ground truth label, heatmap of the uncertainty values.}
        \label{subfig:1}
        \vspace{10pt}
    \end{subfigure}
    \hfill
    \begin{subfigure}[b]{1\textwidth}
        \centering
        \includegraphics[width=\textwidth]{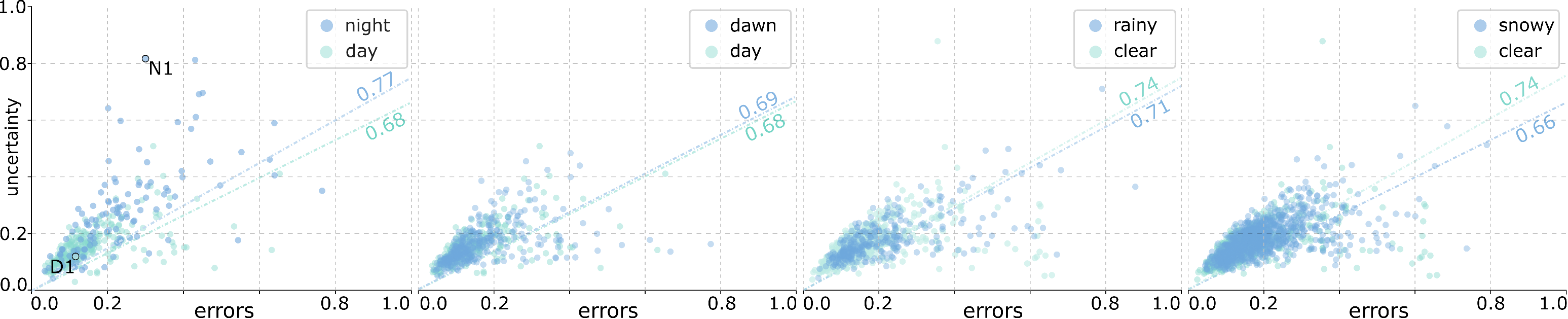}
        \caption{Correlation plots between the error and the uncertainty on the source (day or clear) and the out-of-domain samples (night, dawn, rainy or snowy).}
        \label{subfig:daynight}
    \end{subfigure}
    \caption{Figure (a) displays randomly selected images of the BDD10K dataset from the day and night domain. Additionally, the resulting prediction of the model, the corresponding ground truth labels, and the heat map of the uncertainty values are depicted. 
    In (b), the correlations between the model error and the computed uncertainty for the following domains are presented: day (source) - night (out-of-domain), day (source) - dawn (out-of-domain), clear (source) - rainy (out-of-domain) and clear (source) - snowy (out-of-domain). In each experiment, the model was trained solely on the respective source domain (day or clear). The uncertainty was calculated using five forward passes and a dropout rate of $0.2$.
    For improved visibility in the plots, the uncertainty values are scaled as follows, the mean error is divided by the mean uncertainty value based on the source domain data, and the ratio is applied as a factor to each uncertainty value.}
    \label{fig:out_of_domain}
\end{figure*}

Our second experiment demonstrates the effectiveness of our proposed performance estimation method in the challenging task of semantic segmentation, where real domain shifts present in the BDD10K dataset are deployed. 
Specifically, we use the domain shifts from day to night, day to dawn, and clear to rainy or clear to snowy.

To evaluate the performance of our model, we train it solely on the respective source domains (day or clear) and add dropout layers during inference to enable the use of the Monte Carlo Dropout method for our proposed performance estimation. After training, the uncertainty for every sample of the source domain and the shifted domain using the Monte Carlo Dropout with a dropout rate of 0.2 and five forward passes per sample is calculated. 
The uncertainty is calculated for each pixel. 

It can be depicted with a heat map (see Fig.~\ref{fig:out_of_domain}) as done here for two randomly selected images, one from the source domain day and the other from the shifted domain night.

On both heat maps (day and night), high uncertainty is present along the edges of the segmented objects and areas. Further, in the night domain areas that experience information loss due to the night domain's characteristics, this is visible in large parts of the sky and the poorly illuminated driveable space. Differing pixel class predictions cause uncertainty increases during multiple forward passes - this provides an example of how a \textit{defensive perception envelope} detects the night sample as out-of-domain (see the proposed system flow overview visualized in Fig.~\ref{fig:Framework}). Based on the quantification of the derived uncertainty value, the system can enact a safety countermeasure.

As the ground truth label for each sample of the source domain and the shifted domain is available, it is possible to calculate the true model error for each sample. Spearman's rank correlation coefficient \citep{Spearman_corr} is calculated over all samples of the source and out-of-domain data to examine the relation between the true model error and our proposed performance estimation. The validation reveals a strong correlation between the model's error and the performance estimation. For the day domain, the correlation coefficient is $0.68$, while for the clear domain, it is $0.74$. These source domain correlation coefficients can further serve as a reverence for the out-of-domain correlations. 

Thereupon, for the out-of-domain data, strong correlations are confirmed: A correlation coefficient for the night domain of $0.77$ and $0.68$ for the dawn domain, as well as $0.71$ for the rainy domain and $0.66$ for the snowy domain, shows that the here proposed performance estimation reflects the model's error.

It is to be highlighted that the here presented technique tends to underestimate the prediction error due to a correlation coefficient below one between the model's error and the uncertainty across all examined domains.

This finding is significant for systems under deployment, as this shows that the novel approach can provide a proxy for model performance without the need for ground truth labels at runtime. Furthermore, our proposed uncertainty estimation method reliably approximates the model's prediction error, particularly on out-of-domain data (as evident in Fig.~\ref{fig:out_of_domain}).

\subsection{Sequential data allows for compute efficient uncertainty estimation}
\label{subsec: experiment rolling MC}
As employed in the experiments above, the vanilla Monte Carlo dropout method can become computationally expensive as it requires multiple forward passes per frame to determine the model's uncertainty on a given sample (see Fig.~\ref{fig:Framework}). 
Under deployment, perception systems are heavily limited by runtime requirements, facing limited computational resources \citep{HafnerBackbone}.

To address this issue, we propose a compute-efficient solution by taking advantage of the sequential nature of the input data - as it is known from a camera stream within autonomous vehicles. Furthermore, assuming that the frame rate of the input data is high enough, the differences between subsequent frames can be neglected. Thus, to obtain a sample's uncertainty and performance estimation, instead of multiple forward passes on each frame, we process subsequent frames by applying rolling forward passes (see Rolling Monte Carlo Dropout in Subsec.~\ref{subsec:RollingMC}). 

The model is again trained on the BDD10K training data set, and the experiments are executed on the 20 provided video sequences of the BDD10K. These video sequences are recorded at different locations with various driving velocities. At the same time, the frame rate is constant at 30 frames per second for each video sequence. In addition to verifying the assumption of the similarity of subsequent frames, it is further evaluated how the frame rate and stride of rolling forward passes affect the model's uncertainty. 
The stride with this defines the number of images that are considered to calculate the uncertainty and correspond to the number of forward passes in the vanilla Monte Carlo Dropout.


\begin{figure}
    \centering
    \includegraphics[width=0.49\textwidth]{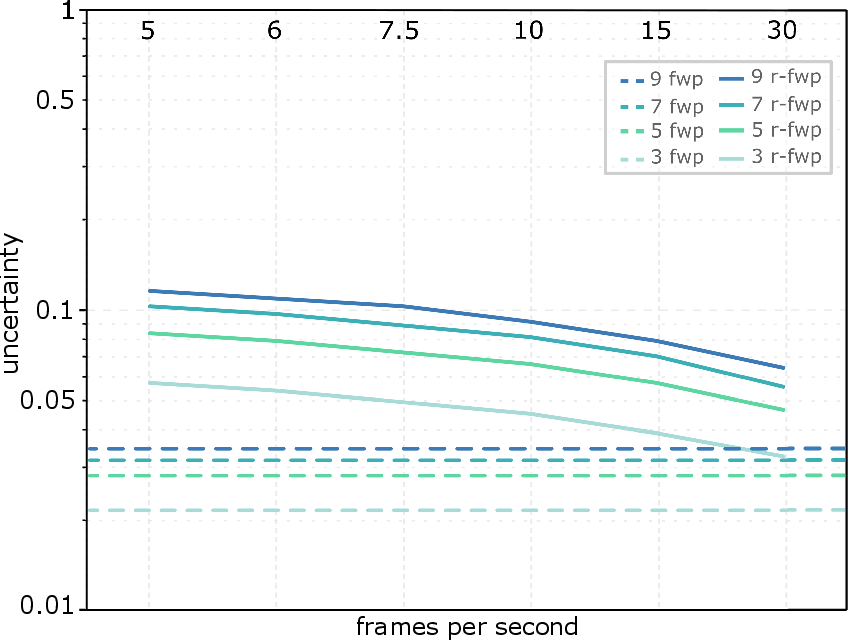}
    \caption{This graph compares the influence of the frame rates on the uncertainty. As a baseline (dotted lines), the frame rate independent Monte Carlo Dropout with a different number of forward passes (fwp) is considered; thus, the uncertainty is calculated on only one image. The Rolling Monte Carlo Dropout (solid lines) applies rolling forward passes (r-fwp) and calculates the uncertainty on subsequent frames, which leads to a frame rate dependency. The model for this graph was trained on the BDD10k training set, and the uncertainty was calculated on the provided 20 video sequences from the BDD10k dataset. A fixed dropout rate of 0.2 is used.}
    \label{fig:fps_slidingwindow}
\end{figure}

\begin{table*}[h!]
    \begin{center}
        \begin{tabular}{c||c|c|c|c|c|c|c|c|c|c}
            dropout rate & 0 & 0.1 & 0.2 & 0.3 & 0.4 & 0.5 & 0.6 & 0.7 & 0.8 & 0.9  \\
            \hline
            error & 0.21 & 0.21 & 0.21 & 0.21 & 0.21 & 0.22 & 0.22 & 0.23 & 0.25 & 0.34 \\
        \end{tabular}
        \caption{The table presents the model's error based on the applied dropout rate, rounded to two significant decimals. When no dropout is applied - the dropout rate is $0$ - the model's error is $0.21$. This error serves as benchmark for comparing the impact of different dropout rates on the model's performance.}
        \label{tab:dropout-table}
    \end{center}
\end{table*}

For 30 frames per second, the highest possible in our experiment, the uncertainty with three forward passes of the vanilla Monte Carlo Dropout is $0.022$. In contrast, for three rolling forward passes, the uncertainty at $30$ frames per second is $0.033$ - a minor deviation of $0.011$ in the uncertainty value (see Fig.~\ref{fig:fps_slidingwindow}). The slight increase in the uncertainty needs an educated trade-off against the advantage of square savings $O(n^2) \to O(n)$ in computing efforts. Accordingly, for the vanilla Monte Carlo dropout, three forward passes must be applied on each of the three consecutive frames, so nine forward passes in total. In contrast, for the Rolling Monte Carlo Dropout, only three forward passes are necessary to cover the three considered frames. 

Furthermore, it is shown that the uncertainty is more sensitive to the stride of the Rolling Monte Carlo Dropout than to the number of forward passes of the vanilla Monte Carlo Dropout. For a larger number of forward passes, the uncertainty slightly increases from $0.032$ (three forward passes) to $0.035$ (nine forward passes). For an increasing stride, the uncertainty increases from $0.033$ (stride of three frames) to $0.065$ (stride of nine frames) and thus almost doubles.

The frame rate only influences the Rolling Monte Carlo Dropout. The frame rate does not affect the vanilla Monte Carlo Dropout, as its uncertainty calculation is based on a single frame. The uncertainty curvature for Rolling Monte Carlo Dropout is characteristically decreasing: The lower the frame rate, the higher the uncertainty as the difference between consecutive frames increases. For a stride of three, the uncertainty increases from $0.032$ up to $0.058$, and for a stride of nine, from an uncertainty of $0.065$ to an uncertainty of $0.118$. 

In conclusion, our results indicate that by applying Rolling Monte Carlo Dropout on consecutive data with a high frame rate, square the computational effort is reduced. This finding is of major importance as computational effort is a very limited resource in autonomous driving systems. Therefore, the following section investigates the effect of the dropout rate on the inference performance.

\subsection{There is no need for a pure inference pass}

Dropout layers are essential for applying the (Rolling) Monte Carlo Dropout to estimate a model's uncertainty during deployment. Using the novel Rolling Monte Carlo Dropout approach, reducing the number of necessary forward passes to only one per frame is possible. As is, another forward pass without dropout is still needed to yield the semantic segmentation output of the perception stack. However, in case the dropout does not reduce the quality of the semantic segmentation during inference, the output of the forward pass with Monte Carlo Dropout can directly be used as the output of the perception stack, which would further half the remaining computational efforts. 

In order to determine whether dropout affects the model's prediction performance, a model is trained on the day domain of the BDD10K training dataset. Subsequently, the model is validated on the validation data set by comparing different dropout rates against the induced error, see Tab. \ref{tab:dropout-table}. 


The results show that up to a dropout rate of 0.4, the error induced by dropout is smaller than $1\%$.
For a dropout rate of up to $0.6$, the error is still around $1\%$. However, the error rises to $0.34$ for a dropout rate of 0.9. For the here presented data set, the dropout rate within a range of 0.1 to 0.6 can be safely used while maintaining the functionality of the semantic segmentation and, at the same time, reducing the computational effort within the \textit{defensive perception envelope}.

\section{Conclusion}
In this paper, we proposed a method for addressing the issue of unnoticed catastrophic deployment and domain shift in neural networks for semantic segmentation in autonomous driving. Our approach is based on the idea that deep learning-based perception for autonomous driving is uncertain and best represented as a probability distribution. Furthermore, we demonstrated the applicability of our method for multiple different potential deployment shifts relevant to autonomous driving, such as entering for the model unknown domains such as night, dawn, rainy, or snowy.

Our \textit{defensive perception envelope} encapsulates the neural network under deployment within an envelope based on the epistemic uncertainty estimation through the Monte Carlo Dropout approach. This approach does not require modification of the deployed neural network and has been shown to guarantee expected model performance. In addition, 
it
estimates a neural network's performance, enabling monitoring and notification of entering domains of reduced neural network performance under deployment.

Furthermore, our envelope is extended by novel methods to improve the application in deployment settings, such as Rolling Monte Carlo Dropout, including reducing compute expenses and confining estimation noise. Finally, by enabling operational design domain recognition via uncertainty, our approach potentially allows for customized defensive perception, safe-state triggers, warning notifications, and feedback for testing or development of the perception stack.

The safety of autonomous vehicles is of paramount importance, and the ability to detect and respond to domain shifts is critical. Our approach shows great potential for application in deployment settings and has the capability to improve the overall safety and performance of autonomous driving systems. By making the source code publicly available, we hope to spark further research in this direction. 

\begin{acknowledgements} 
    We want to thank our fellow researchers at Karlsruhe Institute of Technology and our colleagues at ZF Friedrichshafen AG - in particular, Dr. Jochen Abhau, and apl. Prof. Dr. Markus Reischl.
\end{acknowledgements}

\bibliography{references}
\end{document}